\documentclass[journal,twoside,web]{ieeecolor}

\usepackage{generic}
\usepackage{cite}
\usepackage{amsmath,amssymb,amsfonts}
\usepackage{array}
\usepackage{booktabs}
\usepackage{graphicx}
\usepackage[labelformat=simple]{subcaption}

\usepackage{multirow}
\usepackage{textcomp}
\usepackage{url}
\usepackage{hyperref}
\hypersetup{hidelinks}

\newcommand{\etal}{\emph{et al.}}
\newcommand{\corejepa}{\mbox{CoRe-WM}}
\newcommand{\frs}{\mbox{FRS}}
\newcommand{\irg}{\mbox{IRG}}
\newcommand{\flr}{\mbox{FLR}}
\newcolumntype{L}[1]{>{\raggedright\arraybackslash}p{#1}}

\def\BibTeX{{\rm B\kern-.05em{\sc i\kern-.025em b}\kern-.08em
    T\kern-.1667em\lower.7ex\hbox{E}\kern-.125emX}}
\def\refname{References}

\begin{document}

\title{\corejepa: Conditioned Response-State World Modeling for Longitudinal Treatment-Response Prediction in Breast DCE-MRI}

\author{
Lin Zhao, Bowen Chen, Qiaohui Gao, Shaowen Wan, Xiang Li, Wei Liu, and Tianming Liu%
\thanks{Lin Zhao, Bowen Chen, and Shaowen Wan are with the Department of Biomedical Engineering, New Jersey Institute of Technology, Newark, NJ 07102, USA.}%
\thanks{Qiaohui Gao is with the College of Engineering, Northeastern University, Boston, MA 02115, USA.}%
\thanks{Xiang Li is with the Department of Radiology, Massachusetts General Hospital, Boston, MA 02114, USA.}%
\thanks{Wei Liu is with the Department of Radiation Oncology, Mayo Clinic, Phoenix, AZ 85054, USA.}%
\thanks{Tianming Liu is with the School of Computing, University of Georgia, Athens, GA 30602, USA.}%
\thanks{Corresponding authors: Lin Zhao (lin.zhao.1@njit.edu) and Tianming Liu (tliu@uga.edu).}%
}
\maketitle

\begin{abstract}
Pathologic complete response (pCR) after neoadjuvant chemotherapy is a key prognostic indicator in aggressive breast cancer, yet existing prediction approaches either rely on pretreatment imaging alone, missing the patient's evolving response, or optimize multi-visit classifiers under direct outcome supervision, coupling representation learning with the clinical endpoint. We formulate longitudinal pCR prediction as a world-modeling problem and propose Conditioned Response-state World Modeling (\corejepa{}), which learns a latent state-transition model that predicts the next imaging state from the observed sequence. The transition function is conditioned on patient clinical profile and assigned therapy through additive shift and feature-wise modulation, enabling patient- and treatment-specific dynamics. A Factorized Response State pathway further decouples treatment-response dynamics from imaging appearance by encoding the structural prior that response manifests primarily in lesion morphology. After pretraining, all parameters are frozen and a linear readout predicts pCR at each clinical decision point. We evaluate \corejepa{} on I-SPY2 under five-fold cross-validation and assess cross-cohort transfer on I-SPY1. \corejepa{} achieves the highest discrimination among all evaluated methods at every prospective decision point, and demonstrates robust cross-cohort transfer. Ablation studies identify patient-profile conditioning and the factorized response pathway as the principal contributors. These results establish world-model-based representation learning as a promising framework for outcome prediction in oncology and beyond.
\end{abstract}

\begin{IEEEkeywords}
Breast cancer, dynamic contrast-enhanced MRI, joint-embedding predictive
architecture, longitudinal imaging, neoadjuvant therapy, pathologic complete
response, world model.
\end{IEEEkeywords}

\section{Introduction}
\label{sec:introduction}

\IEEEPARstart{P}{athologic} complete response (pCR), conventionally determined by pathological examination of post-treatment tissue, is strongly associated with favorable long-term outcomes, particularly in aggressive breast cancer subtypes \cite{cortazar2014,esserman2012,yee2020}. Pretreatment prediction of pCR could indicate whether a patient is likely to respond to the planned regimen. Extending such prediction across multiple candidate regimens would further enable estimating and comparing expected response to each option, supporting individualized treatment selection.

Recent studies have investigated pCR prediction in breast cancer using dynamic contrast-enhanced (DCE) MRI, which characterizes contrast-enhancement kinetics related to tumor perfusion, vascular permeability, and angiogenic activity. Early approaches extracted response-related patterns from pretreatment imaging alone \cite{hylton2012,hylton2016,comes2021,nishizawa2025}. Ha \etal{} applied a pretrained CNN to pretreatment postcontrast MRI to predict treatment response \cite{ha2019}, while Braman \etal{} modeled intratumoral and peritumoral DCE-MRI features with multi-institutional validation in patients receiving HER2-targeted therapy \cite{braman2020}. Subsequent work incorporated richer imaging representations and non-imaging variables. Peng \etal{} showed that deep features from pretreatment DCE-MRI provided stronger pCR discrimination than radiomics \cite{peng2022}, and Duanmu \etal{} integrated MRI with molecular and demographic variables to complement baseline imaging \cite{duanmu2020}. Notably, pCR rates vary substantially with patient profiles, including receptor subtype, exceeding 40\% in HER2-positive and triple-negative tumors while remaining below 20\% in HR-positive/HER2-negative disease~\cite{cortazar2014,spring2020,ispy2data}. Receptor status, genomic risk, and other clinical characteristics therefore constitute strong baseline prognostic factors whose interaction with the administered therapy shapes individual treatment response. Across existing imaging methods, these clinical and treatment variables are either omitted or appended as covariates in the outcome classifier rather than integrated into the modeling of treatment response, limiting the capacity to represent patient-specific response patterns.

Beyond this representational gap, these pretreatment approaches cannot incorporate the patient's evolving response once treatment begins or update predictions as new imaging evidence becomes available. As treatment proceeds, serial DCE-MRI examinations reveal patient-specific changes in tumor size, morphology, and enhancement that reflect the ongoing treatment effect \cite{hylton2016}. These on-treatment observations motivate longitudinal approaches that update the pCR estimate as new examinations become available. Several studies have explored this setting. Jing \etal{} used DCE-MRI acquired at multiple treatment timepoints for pCR prediction \cite{jing2024}, Gao \etal{} combined longitudinal imaging with multimodal clinical information \cite{gao2024}, and Tang \etal{} modeled spatiotemporal interactions between baseline and early-treatment MRI \cite{tang2025}. Together, these studies demonstrated that on-treatment imaging provides additional predictive information beyond the pretreatment examination. However, existing longitudinal approaches generally optimize a multi-visit classifier under direct pCR supervision rather than model the dynamics of treatment response. The temporal evolution of tumor state, characterized by the rate, timing, and pattern of morphological change under therapy, carries predictive information distinct from any single-visit or fixed-pair snapshot. Capturing these dynamics requires a model that represents treatment response as an evolving state trajectory and progressively refines its prediction as each new examination arrives.

Motivated by this need, we formulate longitudinal pCR prediction as a world-modeling problem. A world model learns to predict future states from the current state and an applied action, building an internal representation of the underlying dynamics rather than mapping observations directly to outcomes \cite{ha2018worldmodels,hafner2020dreamer,hafner2023dreamerv3,assran2023,bardes2024vjepa}. The world we are modeling is patients' treatment response. As therapy is administered, the lesion evolves, and serial DCE-MRI examinations capture successive snapshots of this evolution. In this formulation, each MRI observation defines the patient state, and the assigned therapy together with individual clinical profiles (e.g., receptor status, age) serve as the condition that governs patient-specific evolution. A learned transition function takes the observed state sequence and this condition to predict the latent representation of the next visit, progressively refining its representation of the treatment response dynamics as each new observation arrives. We instantiate this formulation within the Joint-Embedding Predictive Architecture (JEPA) \cite{assran2023,bardes2024vjepa} as Conditioned Response-state World Modeling (\corejepa). A conditioned causal Transformer forms the transition core, with treatment and clinical information entering through additive shift and feature-wise modulation to shape the dynamics. Because treatment response manifests most directly in changes in lesion volume, shape, and morphology, a parallel Factorized Response State (\frs) pathway encodes this structural prior by operating on compact geometry descriptors, guiding the learned representation toward treatment-relevant changes. Image-derived Response Guidance (\irg) reinforces this pathway through supervision derived from inter-visit morphological and enhancement changes. After pretraining, all parameters are frozen and a Frozen Linear Readout (\flr) maps the accumulated response states to pCR prediction at each decision point. We pretrain and evaluate \corejepa{} on 808 I-SPY2 patients~\cite{ispy2data} with four-visit DCE-MRI under five-fold cross-validation, and assess cross-cohort transfer on 156 I-SPY1 patients~\cite{ispy1data}. \corejepa{} achieves the highest AUROC among all evaluated methods at every prospective decision point, outperforming matched supervised baselines and reproduced longitudinal approaches on I-SPY2, and demonstrates competitive cross-cohort transfer to I-SPY1.

Our contributions are as follows:

a) We formulate longitudinal treatment-response prediction as a world-modeling problem: instead of training a multi-visit classifier that maps imaging directly to pCR, the model learns a state-transition function that predicts the next latent patient state given the current state, assigned therapy, and patient clinical profiles. This world-model formulation shifts the learning objective from outcome classification to modeling treatment-response dynamics, enabling progressive risk updating as new examinations arrive.

b) We introduce a Factorized Response State that decouples treatment-response dynamics from high-dimensional imaging appearance via a compact parallel pathway operating on lesion geometry descriptors. This factorization encodes an explicit structural prior that treatment response manifests primarily in lesion morphology, reducing the representational burden on the world model and its dependence on large training sets.

c) We integrate clinical/treatment conditioning directly into the state-transition function through dual mechanisms (additive shift and feature-wise modulation), so that treatment and patient clinical profiles shape the learned response trajectory rather than entering only at classification time. The condition-permutation analysis confirms that this conditioning is encoded in the frozen response state itself.

d) We evaluate on I-SPY2 under five-fold cross-validation and on I-SPY1 for cross-cohort transfer. \corejepa{} achieves the highest AUROC at every prospective decision point on I-SPY2 and demonstrates cross-cohort robustness on I-SPY1, providing a novel pathway toward world-model-based longitudinal treatment-response prediction.

\section{Related Work}
\label{sec:related_work}

\subsection{Breast DCE-MRI and pCR Prediction}

Quantitative DCE-MRI established that longitudinal changes in functional tumor volume are associated with pathologic response and recurrence-free survival in the I-SPY trials \cite{hylton2012,hylton2016}. These foundational studies demonstrated the prognostic value of serial imaging but relied on handcrafted volumetric endpoints rather than learned representations. Subsequent deep-learning approaches focused on pretreatment imaging: Ha \etal{} used pretrained CNN features from postcontrast MRI \cite{ha2019}, Braman \etal{} modeled intratumoral and peritumoral DCE-MRI features with multi-institutional validation \cite{braman2020}, and Comes \etal{} applied transfer learning to pretreatment scans \cite{comes2021}. Multimodal methods further integrated MRI with molecular and demographic variables \cite{duanmu2020,peng2022}, and attention-based architectures fused whole-breast 3-D MRI with clinical features through self-attention \cite{nishizawa2025}. BreastDCEDL standardized a large multicenter DCE-MRI resource for systematic comparison \cite{breastdcedl2026}. Across these studies, clinical and treatment variables are either absent from the model or appended as covariates in the outcome classifier. \corejepa{} instead integrates these variables directly into a learned state-transition function, enabling treatment-conditioned dynamics modeling that updates predictions as each new visit arrives.

\subsection{Longitudinal Medical Image Modeling}

Longitudinal imaging methods span pairwise change models, sequence encoders, and future-state generators. Siamese and multitask networks compare repeated scans for cancer assessment \cite{veasey2020,jin2021}, while recurrent and attention models aggregate longer imaging histories in neuroimaging and ophthalmology \cite{cui2019,chen2024longformer,holste2024}. More recently, Transformers have been applied to cross-timepoint interactions for treatment response prediction \cite{sun2024lomiat}, and neural ODEs to continuous-time latent disease progression \cite{zeghlache2024latim}. In breast cancer, baseline and on-treatment MRI have been combined through phase concatenation \cite{qu2020}, uncertainty-aware prediction \cite{jing2024}, multimodal longitudinal fusion \cite{gao2024}, and spatiotemporal interaction modeling \cite{tang2025}. Huang \etal{} introduced self-supervised temporal-order pretraining on longitudinal breast MRI with multicenter validation \cite{huang2026bstnet}. More explicit trajectory methods transfer later latent features to an early predictor \cite{jingwang2025}, regularize four-visit phenotype trajectories with responder-alignment objectives \cite{janickova2025}, or construct spatiotemporal response graphs from longitudinal imaging \cite{kiechle2026}. These approaches demonstrate that temporal structure carries information beyond baseline imaging. Most, however, optimize the temporal representation directly for the clinical endpoint, coupling the learning of state dynamics with outcome prediction. This design does not separate the representation of imaging-state evolution from the task of predicting pCR, and may require examinations that are unavailable at earlier decision points. \corejepa{} addresses this by learning from every valid next-visit transition without outcome labels, supporting evaluation at arbitrary observed prefixes, and freezing the trajectory representation before outcome prediction.

\subsection{World Models and JEPA}

World models learn an internal state and dynamics that predict future observations under actions or context. Recurrent visual world models \cite{ha2018worldmodels}, Dreamer \cite{hafner2020dreamer}, and DreamerV3 \cite{hafner2023dreamerv3} demonstrated this paradigm in reinforcement learning, where generative models predict pixels or observations to support planning. JEPA offers a complementary formulation that predicts target representations instead of raw observations: I-JEPA applies this to masked image regions \cite{assran2023}, V-JEPA extends it to video \cite{bardes2024vjepa}, and LeWorldModel studies stable latent prediction with distribution regularization \cite{leworldmodel2026}. Medical adaptations of these ideas are emerging. CheXWorld applied a world-model objective to chest X-ray representation learning \cite{yue2025chexworld}. Volumetric JEPA pretraining has been explored for 3-D medical images \cite{kondepudi2026voljepa}, alongside multimodal image-record prediction \cite{li2025multimodaljepa}, generative tumor simulation for treatment planning \cite{yang2025mewm}, event-based patient simulation from electronic health records \cite{mu2026ehrworld}, and action-conditioned physiologic forecasting \cite{wang2026chronomedicalworld}. These studies collectively demonstrate the potential of predictive state representations in medicine. \corejepa{} extends this direction to longitudinal breast DCE-MRI, using assigned therapy and patient biology as observed context, predicting the next-visit latent rather than future pixels, and coupling imaging appearance with a factorized response state that explicitly tracks treatment-induced morphological change.

\section{Methods}

\begin{figure*}[!t]
\centering
\includegraphics[width=\textwidth]{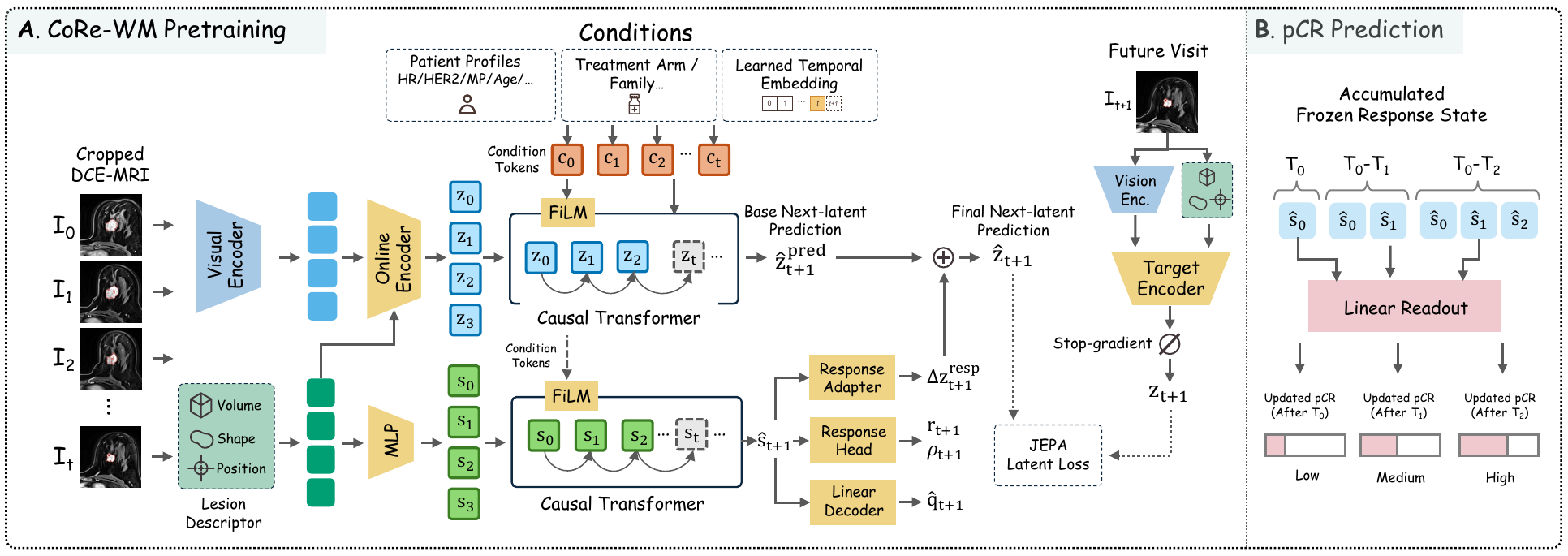}
\caption{Overview of \corejepa. \textbf{(A)}~Outcome-free pretraining. The upper pathway encodes DCE-MRI visits into latent states and predicts the next latent via a conditioned causal Transformer; the lower pathway operates on lesion geometry descriptors to produce a factorized response state $\hat{\mathbf{s}}_{t+1}$, which corrects the latent prediction and is supervised by the \irg{} loss. Patient profile, treatment, and temporal embeddings condition both pathways through FiLM modulation~\cite{perez2018film}. The EMA target encoder~\cite{grill2020} provides the stable prediction target. \textbf{(B)}~pCR prediction. All parameters are frozen; accumulated response states at each decision point are passed to a linear readout for progressively updated pCR estimates.}
\label{fig:architecture}
\end{figure*}

\subsection{Problem Formulation}

We formulate treatment response prediction as a two-stage problem. In the first stage, we learn a conditioned state-transition model that implicitly encodes the patient's treatment response in a self-supervised manner, without access to outcome labels. In the second stage, the learned state representations are frozen and a linear readout predicts clinical outcome from the accumulated state trajectory (Fig.~\ref{fig:architecture}).

At each timepoint $t$, we define the patient's \emph{latent state} $\mathbf{z}_t$ as a joint encoding of the imaging appearance, captured by a volumetric image $I_t$, and the lesion geometry, summarized by a compact descriptor $\mathbf{q}_t \in \mathbb{R}^{D_q}$ (e.g., volume, shape, and position). The state is obtained through a learned encoder $\mathbf{z}_t = \text{Enc}_\theta(I_t, \mathbf{q}_t) \in \mathbb{R}^{D_z}$. The evolution of $\mathbf{z}_t$ across timepoints constitutes the patient's treatment-response trajectory. This trajectory is governed by a \emph{state transition} conditioned on the patient's baseline clinical profile $\mathbf{b}$ (e.g., receptor status) and the administered treatment $\mathbf{a}$ (e.g., drug regimen), denoted collectively as $\mathbf{c} = (\mathbf{b}, \mathbf{a})$. The central task is to learn a transition function
\begin{equation}
  \hat{\mathbf{z}}_{t+1} = T_\theta(\mathbf{z}_{0:t},\; \mathbf{c})
\end{equation}
that, given the observed state prefix $\mathbf{z}_{0:t}$ and the condition $\mathbf{c}$, predicts the next state $\hat{\mathbf{z}}_{t+1}$ in latent space.

In parallel with the image-level prediction, a factorized response pathway (Section~\ref{sec:response_state}) processes the lesion geometry sequence $\mathbf{q}_{0:t}$ under the same condition $\mathbf{c}$ and produces a compact response state $\hat{\mathbf{s}}_{t+1} \in \mathbb{R}^{D_s}$ ($D_s \ll D_z$) that captures treatment-response dynamics in a low-dimensional space separate from imaging appearance. Once the transition model is trained, all parameters are frozen. For a patient with $T$ observed visits, a linear readout aggregates the accumulated response states $\hat{\mathbf{s}}_1, \ldots, \hat{\mathbf{s}}_{T-1}$ and maps them to the binary endpoint $y \in \{0, 1\}$.

\subsection{Latent State Encoder}

Each visit is characterized by a volumetric image $I_t$ and a compact lesion descriptor $\mathbf{q}_t \in \mathbb{R}^{D_q}$ that summarizes the geometry of the region of interest within the imaging field. We encode these two modalities separately and combine them additively:
\begin{equation}
  \mathbf{z}_t = g_\theta\!\bigl(f_\theta(I_t)\bigr) + h_\theta(\mathbf{q}_t)
\end{equation}
where $f_\theta$ is a 3-D residual convolutional network that maps the volumetric input to an appearance vector, $g_\theta$ is a nonlinear projector, and $h_\theta$ is a geometry projector. All three operate in a shared $D_z$-dimensional latent space. The appearance encoder progressively downsamples the input volume through multiple residual stages to a single vector, which is then projected into the latent space. The geometry projector maps the low-dimensional descriptor $\mathbf{q}_t$ into the same space through a lightweight MLP.

The encoder above, referred to as the \emph{online encoder}, encodes the observed states that serve as input to the transition function. A second encoder, the \emph{target encoder}, provides the prediction targets representing future states. The target encoder shares the same architecture but maintains separate parameters $\phi$, updated as an exponential moving average (EMA) of the online parameters: $\phi \leftarrow \mu\,\phi + (1 - \mu)\,\theta$ after each gradient step. Its slowly evolving parameters provide a stable prediction objective that prevents degenerate solutions such as representational collapse. This online-target pair forms the basis of the JEPA~\cite{assran2023,bardes2024vjepa} framework adopted in our transition function.

\subsection{Conditioned State Transition}

The clinical profile $\mathbf{b}$ and treatment $\mathbf{a}$ are each encoded through learned embeddings for categorical variables and normalization for continuous variables, then concatenated to form $\mathbf{c} \in \mathbb{R}^{D_c}$. Since $\mathbf{c}$ is shared across timepoints, we concatenate a learned temporal embedding at each step to yield per-step condition vectors $\mathbf{c}_t \in \mathbb{R}^{D_c + D_\tau}$.

Given the observed latent state sequence $(\mathbf{z}_0, \ldots, \mathbf{z}_t)$ and the corresponding condition sequence $(\mathbf{c}_0, \ldots, \mathbf{c}_t)$, the conditioned state transition estimates the next latent state. We adopt a conditioned causal Transformer as the transition core. The Transformer encoder uses pre-normalization, an upper-triangular causal mask ensuring that the output at position $t$ attends only to positions $0, \ldots, t$, and learned positional embeddings to encode temporal position. The condition vector $\mathbf{c}_t$ enters through two complementary mechanisms:

\paragraph{Additive conditioning.} A linear projection maps $\mathbf{c}_t$ to $D_z$ dimensions and adds it to the input token at each position before the Transformer layers, providing a global context shift.

\paragraph{Feature-wise linear modulation (FiLM).} Within each Transformer layer, a learned projection of $\mathbf{c}_t$ produces per-dimension scale and shift parameters $(\gamma_t, \beta_t) \in \mathbb{R}^{D_z}$, which modulate the layer's hidden state $\mathbf{h}_t$ following the FiLM conditioning mechanism~\cite{perez2018film}:
\begin{equation}
  \tilde{\mathbf{h}}_t = \mathbf{h}_t \odot \bigl(1 + \eta\,\tanh(\gamma_t)\bigr) + \eta\,\beta_t
\end{equation}
where $\eta$ is a fixed scaling factor and $\odot$ denotes element-wise multiplication. The dual conditioning mechanism balances two competing requirements: the transition should be aware of the treatment context, which determines the expected biological response, but should not learn to predict outcomes from clinical features alone. To this end, the FiLM weights and biases are initialized to zero, so conditioning begins as a near-identity transformation and grows in influence only as training progresses, ensuring that the state dynamics remain the primary signal. A prediction head projects the Transformer output to the latent prediction $\hat{\mathbf{z}}^{\text{pred}}_{t+1}$.

\subsection{Factorized Future Response State}\label{sec:response_state}

While the conditioned state transition captures the visual evolution, the high dimensionality of volumetric latents ($D_z$) mixes appearance, contrast, and geometry information. We introduce a parallel pathway that operates on the compact lesion descriptors $\mathbf{q}_{0:t}$ and produces a lower-dimensional future response state $\hat{\mathbf{s}}_{t+1} \in \mathbb{R}^{D_s}$ ($D_s \ll D_z$) that explicitly tracks the treatment-response trajectory. This factorization encodes the structural prior that treatment response manifests primarily in lesion morphology, guiding the learned representation toward treatment-relevant changes and producing a compact representation suitable for downstream linear readout.

\subsubsection{Temporal Response Dynamics}

The observed lesion descriptors are first log-transformed on volume-related features to stabilize scale differences across patients, then projected through a two-layer network to a $D_s$-dimensional response embedding. A dedicated conditioned causal Transformer, sharing the same FiLM conditioning architecture as the conditioned state transition but with fewer layers, processes the response sequence. The Transformer output at position $t$ defines the future response state $\hat{\mathbf{s}}_{t+1} \in \mathbb{R}^{D_s}$. A linear decoder head maps $\hat{\mathbf{s}}_{t+1}$ back to the descriptor space, producing a geometry estimate $\hat{\mathbf{q}}_{t+1}$. A separate response head decodes a predicted response vector $\hat{\mathbf{r}}_{t+1} \in \mathbb{R}^{D_r}$ and a scalar response score $\hat{\rho}_{t+1}$ from $\hat{\mathbf{s}}_{t+1}$.

These predictions are supervised by an Image-derived Response Guidance (\irg) loss $\mathcal{L}_{\text{IRG}}$ constructed entirely from observed imaging, without reference to the clinical outcome $y$. At each transition, two outcome-free targets are extracted: a response vector $\mathbf{r}_{t+1}$ that concatenates baseline-relative and previous-visit-relative changes in lesion volume, shape, and enhancement intensity, and a scalar response score $\rho_{t+1}$ that aggregates key response components with treatment-category-level standardization. The loss is
\begin{align}
  \mathcal{L}_{\text{IRG}} = \;&\ell_{\text{H}}(\hat{\mathbf{q}}_{t+1},\, \mathbf{q}_{t+1})
    + \ell_{\text{H}}(\hat{\mathbf{r}}_{t+1},\, \mathbf{r}_{t+1}) \nonumber \\
    &+ \lambda_{\text{rank}}\!\!\!\sum_{(i,j):\,\rho_i>\rho_j}\!\!\max\!\bigl(0,\,m - (\hat{\rho}_i - \hat{\rho}_j)\bigr) \nonumber \\
    &+ \lambda_{\text{ctr}} \mathcal{L}_{\text{ctr}}(\hat{\mathbf{s}},\, \mathbf{r})
\end{align}
where $\ell_{\text{H}}$ denotes smooth-$L_1$ loss, the third term is a pairwise margin ranking loss that preserves the ordinal structure of response magnitude across patients within each batch, and $\mathcal{L}_{\text{ctr}}$ is a contrastive loss that pulls response states of patients with similar imaging changes together in the latent space. The geometry estimate $\hat{\mathbf{q}}_{t+1}$ also serves as input to the latent correction described below.

\subsubsection{Latent Correction}

The response pathway contributes a learned correction to the conditioned state transition through a latent adapter. The adapter concatenates the future response state $\hat{\mathbf{s}}_{t+1}$ with the geometry estimate $\hat{\mathbf{q}}_{t+1}$ produced by the response Transformer and projects this to the image latent dimension:
\begin{equation}
  \Delta\mathbf{z}^{\text{resp}}_{t+1} = \alpha_z \cdot \text{Adapt}\!\bigl([\hat{\mathbf{s}}_{t+1};\, \hat{\mathbf{q}}_{t+1}]\bigr)
\end{equation}
with a small scaling factor $\alpha_z$ and zero-initialized output weights. This ensures the response correction is negligible at the start of training, preventing the response pathway from interfering with the conditioned state transition before either component has learned useful representations. As training proceeds, the adapter learns to communicate response-state information to the image prediction in a complementary fashion.

\subsection{Joint Prediction and Pretraining Objective}

The final predicted next-visit latent is the sum of the two pathway outputs:
\begin{equation}
  \hat{\mathbf{z}}_{t+1} = \hat{\mathbf{z}}^{\text{pred}}_{t+1} + \Delta\mathbf{z}^{\text{resp}}_{t+1}
\end{equation}
and the prediction target is the EMA-encoded visit state:
\begin{equation}
  \mathbf{z}^{\text{tar}}_{t+1} = g_\phi\!\bigl(f_\phi(I_{t+1})\bigr) + h_\phi(\mathbf{q}_{t+1}).
\end{equation}
Because the target includes both appearance and geometry, the joint prediction accounts for both modalities. The state transition carries the dominant appearance prediction, and the response-state correction allows the factorized response pathway to refine the forecast when lesion-dynamics information provides complementary evidence.

The total pretraining loss is a weighted combination of three groups of terms:
\begin{equation}
  \mathcal{L} = \lambda_{\text{pred}} \mathcal{L}_{\text{pred}} + \lambda_{\text{sig}} \mathcal{L}_{\text{SIGReg}} + \mathcal{L}_{\text{IRG}}.
\end{equation}
The primary signal $\mathcal{L}_{\text{pred}}$ is the layer-normalized mean squared error between predicted and target latents, summed over all $T{-}1$ transitions with a decaying weight schedule that emphasizes earlier steps. A distributional regularizer $\mathcal{L}_{\text{SIGReg}}$ (SIGReg~\cite{leworldmodel2026}) prevents representational collapse by encouraging the latent distribution to remain well-spread.

The imaging response guidance $\mathcal{L}_{\text{IRG}}$, defined in Section~\ref{sec:response_state}, supervises the factorized response pathway using outcome-free targets derived from inter-visit changes in lesion geometry and imaging intensity.

\subsection{Frozen Linear Readout}

After pretraining, all model weights are frozen and outcome labels $y$ are introduced exclusively at this stage. At each decision point, visits up to timepoint $k$ have been observed. The response pathway processes these visits and produces future response states $\hat{\mathbf{s}}_1, \ldots, \hat{\mathbf{s}}_{k+1}$, from which a feature vector is constructed by concatenating summary statistics of the observed prefix (initial state, current state, prefix mean, recent revision, cumulative displacement). A single class-balanced logistic regression is trained on the pooled features from all patients and all decision points to predict $y$. As more visits are observed, the readout receives a longer state prefix, allowing the prediction to be progressively refined with additional treatment-response information.

\section{Experiments}
\label{sec:experiments}

\subsection{Datasets and Evaluation Protocol}
\label{sec:datasets}

\textbf{Study cohorts.} We used DCE-MRI and clinical data from the I-SPY2 (808 patients, 275 pCR) and I-SPY1 (156 patients, 41 pCR) neoadjuvant breast cancer trials \cite{ispy2data,ispy1data}. Both trials acquired MRI at four treatment timepoints (T0--T3). Prediction was evaluated at T0, T0--T1, and T0--T2; T3 was used only during outcome-free pretraining.

\textbf{DCE-MRI preprocessing.} Each visit was represented by an eight-channel volume comprising precontrast, early postcontrast, late postcontrast, two subtraction, peak enhancement, washout, and binary lesion-region channels. Phase selection followed the BreastDCEDL adaptive protocol \cite{breastdcedl2026}. Lesion regions were derived from the released functional-tumor-volume analysis masks; automatic enhancement-based localization was used when a compatible mask was unavailable. All image channels were robustly normalized per visit, and each volume was cropped or padded to $32\times96\times96$ voxels centered on the T0 lesion. A nine-dimensional geometry descriptor was extracted from each ROI mask.

\textbf{Five-fold evaluation.} Five disjoint patient-level folds were created, stratified by pCR status and HR/HER2 subtype. In each fold, approximately 525 I-SPY2 patients were used for training, 121 for validation, and 162 for testing; the same fold manifest was used for all methods. Reported metrics are fold mean $\pm$ standard deviation, quantifying partition variability. Within each fold, the four non-test partitions were split into training and validation sets; hyperparameters, checkpoints, and classification thresholds were selected on the validation set.

\textbf{Cross-cohort transfer.} Each source-fold model, pCR readout, and operating threshold was fitted using I-SPY2 only, and its probability was computed for every I-SPY1 patient. The five source-fold probabilities were averaged for each patient, and the mean source-validation threshold was applied without using I-SPY1 outcomes.

\textbf{Ethics statement.} Both I-SPY1 and I-SPY2 are multi-center clinical trials approved by the institutional review boards at each participating site; all participants provided written informed consent prior to enrollment. The imaging and clinical data used in this study were obtained from publicly available, de-identified repositories \cite{ispy1data,ispy2data}. No additional ethics approval was required for this secondary analysis of de-identified data.

\subsection{Implementation Details}
\label{sec:implementation}

\textbf{Network configuration.} The 3-D encoder used four residual stages (16, 32, 64, 128 channels), producing a 192-dimensional visit representation via global pooling and layer normalization. The transition predictor had three Transformer blocks (4 heads, 512-dim feed-forward, dropout 0.1) conditioned on a 25-dimensional vector encoding visit position, treatment, molecular profile, and age. The factorized response branch mapped the nine-dimensional lesion descriptor to a 64-dimensional response state.

\textbf{Optimization.} Models were trained with AdamW (lr $5\times10^{-5}$, weight decay $10^{-4}$, batch size 32) for at most 100 epochs with early stopping (patience 4). The target encoder used EMA momentum 0.996. Future-transition losses for T1, T2, and T3 were weighted 2, 1, and 0.5. Loss weights were $\lambda_{\text{pred}} = 1.0$, $\lambda_{\text{sig}} = 0.09$, $\lambda_{\text{rank}} = 1.0$ (margin $m = 0.1$), $\lambda_{\text{ctr}} = 0.02$. The FiLM scaling factor was $\eta = 0.1$, and the response adapter scaling factor was $\alpha_z = 0.05$.

\textbf{Readout and metrics.} After pretraining, all parameters were frozen and a class-balanced logistic readout was fitted on frozen response states; penalty type and regularization strength were selected on the validation fold. AUROC was the primary metric; thresholds were selected by maximizing Youden's index on validation data. Confidence intervals used patient-level bootstrap of pooled out-of-fold predictions.

\subsection{Comparison with Other Methods}
\label{sec:main_comparison}

\subsubsection{Compared Methods}

We compared \corejepa{} with published methods and matched baselines under the same five folds. For pretreatment methods, \textbf{ViT}~\cite{breastdcedl2026} used the BreastDCEDL MinCrop representation with a ViT-B/16 initialized from ImageNet masked-autoencoder weights, evaluated both as a standalone image encoder and with its full two-stage clinical--geometry fusion pipeline. \textbf{3-D ResNet} was a matched pretreatment baseline that encoded the T0 eight-channel volume with the same four-stage 3-D residual network (16, 32, 64, 128 channels) used in \corejepa{} and mapped its globally pooled representation directly to pCR. For longitudinal methods, \textbf{PrefixGRU} was a matched sequential baseline that applied a shared 3-D encoder to each available visit and aggregated the ordered visit representations with a GRU before pCR classification. \textbf{Temporal representation learning}~\cite{janickova2025} used MIP-based inputs with reconstruction, temporal-continuity, and responder-alignment objectives. The \textbf{two-stage dual-task} model of Jing and Wang~\cite{jingwang2025} jointly optimized pCR prediction and future-latent prediction from three-channel early-subtraction inputs with clinical variables. \textbf{BSTNet}~\cite{huang2026bstnet} used temporal-order pretraining on peak-postcontrast volumes from all four visits; the pretrained encoder was retained and a lightweight pCR head was retrained to accept T0--T2 prefixes. Reimplementations followed released architectures when code was available and were selected without test-fold data.

\subsubsection{Results}

\begin{table*}[!t]
\centering
\caption{Five-fold I-SPY2 comparison using the same patient splits and seed for all methods. Values are fold mean $\pm$ standard deviation. The ``Available MRI'' column lists the examinations used by the pCR predictor; T3 may additionally be used as an outcome-free pretraining target (\corejepa{} and BSTNet temporal-order pretraining). ViT two-stage fusion uses the complete BreastDCEDL clinical and geometry feature set. Accuracy, sensitivity, and specificity use a validation-selected threshold.}
\label{tab:main_5fold_comparison}
\scriptsize
\setlength{\tabcolsep}{4pt}
\renewcommand{\arraystretch}{1.07}
\begin{tabular}{@{}p{1.2cm} l c c c c c c@{}}
\toprule
& Method & Avail.\ MRI & AUROC & AUPRC & Acc. & Sens. & Spec. \\
\midrule
\multirow{3}{*}{\shortstack{\emph{Pretreatment}\\\emph{methods}}}
& ViT~\cite{breastdcedl2026}
& T0 & $0.5591\pm0.0320$ & $0.4064\pm0.0156$ & $0.5657\pm0.0507$ & $0.5164\pm0.1935$ & $0.5917\pm0.1741$ \\
& ViT + two-stage fusion~\cite{breastdcedl2026}
& T0 & $0.5950\pm0.0130$ & $0.4230\pm0.0194$ & $0.5521\pm0.0629$ & $0.5745\pm0.2431$ & $0.5412\pm0.2194$ \\
& 3-D ResNet
& T0 & $0.5444\pm0.0391$ & $0.3685\pm0.0242$ & $0.4977\pm0.0617$ & $0.6836\pm0.1546$ & $0.4022\pm0.1661$ \\
\midrule
\multirow{4}{*}{\shortstack{\emph{Longitudinal}\\\emph{methods}}}
& PrefixGRU
& T0--T2 & $0.5784\pm0.0668$ & $0.4157\pm0.0799$ & $0.5718\pm0.0514$ & $0.4982\pm0.2075$ & $0.6098\pm0.1709$ \\
& Temp.\ repr.\ learning~\cite{janickova2025}
& T0--T2 & $0.5179\pm0.0516$ & $0.3710\pm0.0559$ & $0.5606\pm0.0476$ & $0.3527\pm0.2079$ & $0.6678\pm0.1748$ \\
& Two-stage dual-task~\cite{jingwang2025}
& T0--T1 & $0.6958\pm0.0193$ & $0.5412\pm0.0378$ & $0.6435\pm0.0323$ & $0.5745\pm0.0539$ & $0.6792\pm0.0600$ \\
& BSTNet~\cite{huang2026bstnet}
& T0--T2 & $0.6578\pm0.0555$ & $0.4773\pm0.0740$ & $0.6052\pm0.0739$ & $0.5964\pm0.1633$ & $0.6095\pm0.1505$ \\
\midrule
\multirow{3}{*}{\shortstack{\emph{Proposed}\\\emph{method}}}
& \corejepa{} + \flr
& T0 & $0.7337\pm0.0320$ & $0.6107\pm0.0474$ & $0.6412\pm0.0588$ & $0.6582\pm0.2075$ & $0.6325\pm0.1860$ \\
& \corejepa{} + \flr
& T0--T1 & $0.7545\pm0.0468$ & $0.6401\pm0.0461$ & $0.6980\pm0.0572$ & $0.6473\pm0.0625$ & $0.7243\pm0.0996$ \\
& \corejepa{} + \flr
& T0--T2 & $0.7620\pm0.0363$ & $0.6381\pm0.0307$ & $0.7141\pm0.0381$ & $0.6436\pm0.1224$ & $0.7504\pm0.1116$ \\
\bottomrule
\end{tabular}
\end{table*}

Table~\ref{tab:main_5fold_comparison} shows that \corejepa{} produced the strongest discrimination at every prospective decision point. At T0, its AUROC of $0.7337\pm0.0320$ exceeded the ViT ($0.5591\pm0.0320$), the two-stage fusion model ($0.5950\pm0.0130$), and the matched 3-D ResNet ($0.5444\pm0.0391$). Once T1 became available, \corejepa{} reached $0.7545\pm0.0468$, compared with $0.6958\pm0.0193$ for the two-stage dual-task model. In pooled out-of-fold predictions, the AUROC difference between these two methods at T0--T1 was 0.0673 (95\% CI, 0.0344--0.0995; $p=0.0004$). The proposed model improved further to $0.7620\pm0.0363$ with T2, whereas PrefixGRU ($0.5784\pm0.0668$), temporal representation learning ($0.5179\pm0.0516$), and BSTNet ($0.6578\pm0.0555$) all remained substantially lower at the same T0--T2 decision point.

The trajectory results also distinguish the contribution of newly observed response from baseline prognostic information. The mean AUROC increased by 0.0208 from T0 to T0--T1 and by a further 0.0075 from T0--T1 to T0--T2, while specificity rose from 0.6325 to 0.7504. The conditioned trajectory adds discriminative information beyond pretreatment phenotype while supporting sequential prediction from a common frozen state.

\subsection{Cross-Cohort Transfer to I-SPY1}
\label{sec:external_transfer}

\begin{table*}[!t]
\centering
\caption{Cross-cohort transfer from I-SPY2 to 156 I-SPY1 patients (41 pCR). AUROC brackets are patient-bootstrap 95\% confidence intervals. All pCR supervision, model selection, and operating-threshold selection use I-SPY2 only.}
\label{tab:ispy1_transfer}
\scriptsize
\setlength{\tabcolsep}{4pt}
\renewcommand{\arraystretch}{1.07}
\begin{tabular}{@{}p{1.2cm} l c c c c c c@{}}
\toprule
& Method & Avail.\ MRI & AUROC (95\% CI) & AUPRC & Acc. & Sens. & Spec. \\
\midrule
\multirow{3}{*}{\shortstack{\emph{Pretreatment}\\\emph{methods}}}
& ViT~\cite{breastdcedl2026}
& T0 & $0.5584\,[0.4549,0.6609]$ & $0.3159$ & $0.4872$ & $0.6829$ & $0.4174$ \\
& ViT + two-stage fusion~\cite{breastdcedl2026}
& T0 & $0.5750\,[0.4764,0.6746]$ & $0.3276$ & $0.4487$ & $0.6585$ & $0.3739$ \\
& 3-D ResNet
& T0 & $0.6738\,[0.5698,0.7709]$ & $0.4213$ & $0.2949$ & $0.9512$ & $0.0609$ \\
\midrule
\multirow{4}{*}{\shortstack{\emph{Longitudinal}\\\emph{methods}}}
& PrefixGRU
& T0--T2 & $0.6494\,[0.5510,0.7428]$ & $0.3798$ & $0.6282$ & $0.6341$ & $0.6261$ \\
& Temp.\ repr.\ learning~\cite{janickova2025}
& T0--T2 & $0.5548\,[0.4519,0.6579]$ & $0.3433$ & $0.7308$ & $0.0732$ & $0.9652$ \\
& Two-stage dual-task~\cite{jingwang2025}
& T0--T1 & $0.7635\,[0.6780,0.8402]$ & $0.5391$ & $0.7372$ & $0.2439$ & $0.9130$ \\
& BSTNet~\cite{huang2026bstnet}
& T0--T2 & $0.4390\,[0.3392,0.5387]$ & $0.2295$ & $0.3846$ & $0.5854$ & $0.3130$ \\
\midrule
\multirow{3}{*}{\shortstack{\emph{Proposed}\\\emph{method}}}
& \corejepa{} + \flr{}
& T0 & $0.7396\,[0.6523,0.8178]$ & $0.4439$ & $0.6667$ & $0.7561$ & $0.6348$ \\
& \corejepa{} + \flr{}
& T0--T1 & $0.7321\,[0.6396,0.8177]$ & $0.4774$ & $0.7179$ & $0.5610$ & $0.7739$ \\
& \corejepa{} + \flr{}
& T0--T2 & $0.7082\,[0.6114,0.7986]$ & $0.4515$ & $0.7179$ & $0.3659$ & $0.8435$ \\
\bottomrule
\end{tabular}
\end{table*}

We examine cross-cohort robustness in I-SPY1 and Table~\ref{tab:ispy1_transfer} shows substantial variation in transfer performance. The two-stage dual-task model achieved the highest early-treatment AUROC at 0.7635, while \corejepa{} achieved 0.7396 at T0 and 0.7321 at T0--T1. The paired AUROC difference between \corejepa{} and the two-stage dual-task model at T0--T1 was $-0.0314$ (95\% CI, $-0.1274$--0.0606; $p=0.5167$). The 3-D ResNet reached 0.6738 at T0, PrefixGRU reached 0.6494 at T0--T2, and BSTNet reached 0.4390 at T0--T2. Temporal representation learning reached 0.5548 at T0--T2. At T0, \corejepa{} exceeded the ViT, its two-stage fusion, and temporal representation learning by 0.1812, 0.1646, and 0.1922 AUROC, respectively ($p=0.0080$, $p=0.0124$, $p=0.0008$). These results indicate that temporal capacity alone did not ensure robustness across trial cohorts. Unlike its I-SPY2 trajectory, \corejepa{} did not improve as later I-SPY1 examinations were added: AUROC decreased from 0.7396 at T0 to 0.7321 at T0--T1 and 0.7082 at T0--T2. The transferred operating points were imbalanced for the two-stage dual-task model (high specificity, low sensitivity), and BSTNet fell to near-chance AUROC (0.4390), indicating that its temporal-order pretraining did not confer cross-cohort robustness. These results support cross-cohort feasibility and robustness of \corejepa{}.

\subsection{Ablation Study}
\label{sec:ablation}

We evaluated the proposed components through a cumulative ablation in which modules were added one at a time to an image-only JEPA. Every configuration used the same five I-SPY2 folds, seed, pretraining budget, and frozen-readout protocol, and was evaluated at the early-treatment T0--T1 decision point. Table~\ref{tab:additive_ablation} reports model-native fold metrics; the AUROC increment in each row is measured relative to the immediately preceding configuration.

\begin{table*}[!t]
\centering
\caption{Cumulative five-fold ablation at T0--T1. Values are fold mean $\pm$ standard deviation. $\Delta$ AUROC is relative to the preceding row. Patient profile includes HR, HER2, MammaPrint, and age; Tx denotes treatment conditioning; Geo denotes ROI geometry. FRS and IRG denote factorized response state and image-derived response guidance, respectively.}
\label{tab:additive_ablation}
\scriptsize
\setlength{\tabcolsep}{2.3pt}
\renewcommand{\arraystretch}{1.07}
\begin{tabular}{L{4.4cm}c c c c c c}
\toprule
Cumulative configuration & Profile & Tx & Geo & FRS/IRG & AUROC & $\Delta$ AUROC \\
\midrule
Image-only JEPA
& -- & -- & -- & -- & $0.5816\pm0.0316$ & -- \\
$+$ patient profile
& $\checkmark$ & -- & -- & -- & $0.7135\pm0.0117$ & $+0.1319\pm0.0350$ \\
$+$ treatment conditioning
& $\checkmark$ & $\checkmark$ & -- & -- & $0.7115\pm0.0393$ & $-0.0019\pm0.0467$ \\
$+$ ROI geometry
& $\checkmark$ & $\checkmark$ & $\checkmark$ & -- & $0.7380\pm0.0138$ & $+0.0264\pm0.0451$ \\
$+$ factorized response state
& $\checkmark$ & $\checkmark$ & $\checkmark$ & FRS & $0.7506\pm0.0473$ & $+0.0126\pm0.0442$ \\
$+$ image-derived response guidance (\corejepa{})
& $\checkmark$ & $\checkmark$ & $\checkmark$ & FRS+IRG & $\mathbf{0.7545\pm0.0468}$ & $+0.0039\pm0.0082$ \\
\bottomrule
\end{tabular}
\end{table*}

The largest gain came from conditioning the image trajectory on patient profile, which increased mean AUROC from 0.5816 to 0.7135. Treatment conditioning did not provide an additional marginal AUROC gain after the clinical variables were introduced, although it remained part of the transition formulation because treatment specifies the context under which the next state is predicted. Adding explicit ROI geometry increased mean AUROC to 0.7380; in pooled out-of-fold predictions, its gain was 0.0253 (95\% CI, 0.0001--0.0507). The factorized response state and image-derived response guidance further increased mean AUROC to 0.7506 and 0.7545, respectively. Overall, the ablation indicates that patient profile provides the dominant conditioning signal, while explicit lesion geometry and imaging-derived response guidance contribute smaller but complementary improvements beyond the image-only trajectory.

\begin{figure*}[!t]
\centering
\begin{subfigure}[t]{0.245\textwidth}
\centering
\includegraphics[width=\textwidth]{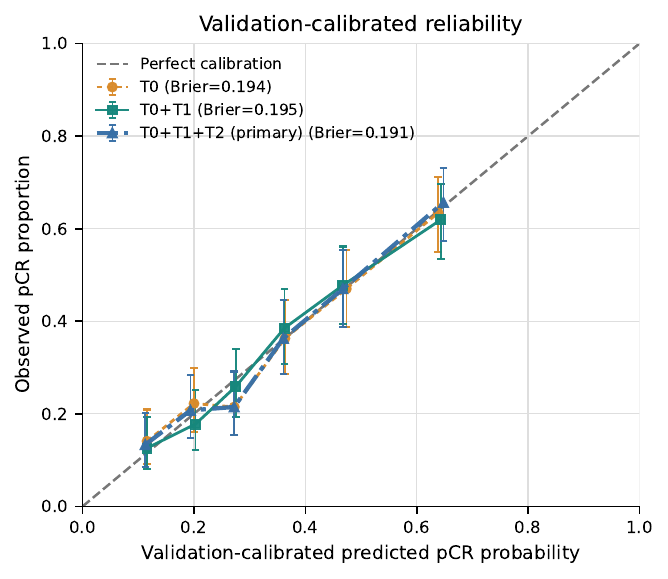}
\caption{}
\label{fig:calibration_5fold}
\end{subfigure}\hfill
\begin{subfigure}[t]{0.245\textwidth}
\centering
\includegraphics[width=\textwidth]{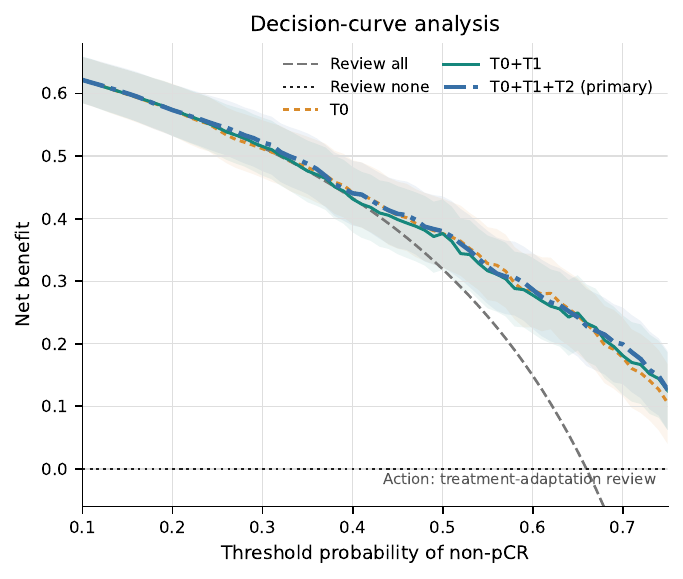}
\caption{}
\label{fig:decision_curve_5fold}
\end{subfigure}\hfill
\begin{subfigure}[t]{0.245\textwidth}
\centering
\includegraphics[width=\textwidth]{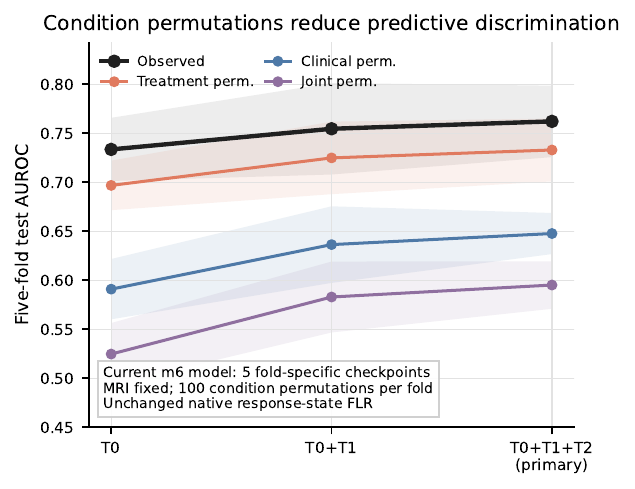}
\caption{}
\label{fig:condition_dependence_5fold}
\end{subfigure}\hfill
\begin{subfigure}[t]{0.245\textwidth}
\centering
\includegraphics[width=\textwidth]{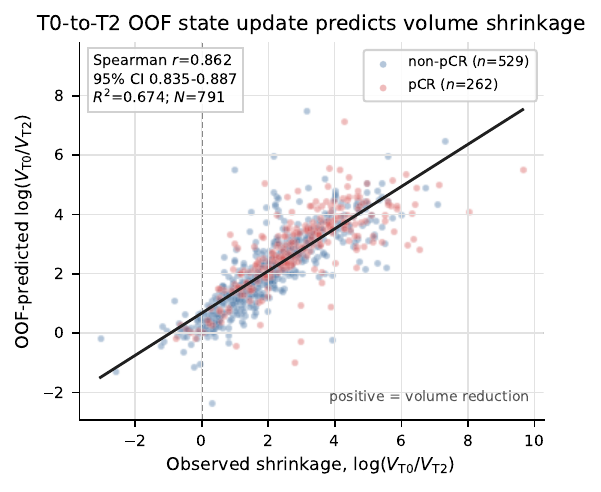}
\caption{}
\label{fig:response_state_grounding_5fold}
\end{subfigure}
\caption{Five-fold diagnostic analyses. (a)~OOF calibration curves at three prospective decision points; points show equal-frequency bins with Wilson 95\% CIs. (b)~Decision-curve analysis for non-pCR identification; shaded regions are bootstrap 95\% CIs. (c)~Condition-dependence analysis; MRI latents and geometry are fixed while treatment, clinical profile, or their joint context is permuted; shaded regions show $\pm1$ SD across folds. (d)~Response-state grounding; a ridge probe maps the frozen T0-to-T2 state update to observed log-volume shrinkage in 791 OOF patients.}
\label{fig:analysis}
\end{figure*}

\subsection{Calibration and Decision-Curve Analysis}
\label{sec:calibration_decision}

Discrimination alone does not establish whether the predicted probabilities are reliable enough for threshold-based use. We therefore evaluated calibration and decision-curve utility from the five held-out I-SPY2 partitions. Within each outer fold, a Platt calibrator was fitted using that fold's validation patients and applied without refitting to its test patients. Concatenating the five test partitions yielded one out-of-fold (OOF) probability for each of the 808 I-SPY2 patients. The resulting pooled OOF AUROC is a different estimand from the arithmetic fold-mean AUROC in Table~\ref{tab:main_5fold_comparison}; in particular, fold-specific calibration can change rankings between patients belonging to different folds while preserving rankings within each fold.

The T0--T2 model had the strongest overall probability performance, with pooled OOF AUROC 0.7288, AUPRC 0.5853, and a calibrated Brier score of 0.1911 (95\% CI, 0.1780--0.2045), compared with calibrated Brier scores of 0.1942 at T0 and 0.1949 at T0--T1. Calibration at T0--T2 was close to the ideal line, with intercept $-0.020$, slope 0.978, and expected calibration error 0.0176; Platt scaling reduced its Brier score from 0.2166 before calibration (Fig.~\ref{fig:calibration_5fold}). These findings indicate that the probabilities were well aligned with observed pCR frequency after validation-only recalibration, while the improvement over the earlier decision points was modest.

For decision-curve analysis \cite{vickers2006}, non-pCR was defined as the actionable event and referral for treatment-adaptation review as the modeled action. At T0--T2, the \corejepa{} curve exceeded both review-all and review-none strategies across probability thresholds from 0.22 to 0.75 (Fig.~\ref{fig:decision_curve_5fold}). This range suggests potential utility when the cost of unnecessary review is traded against the cost of missing a patient unlikely to achieve pCR. However, confidence bands overlapped across decision points, and decision-curve net benefit is calculated from predictive outcomes rather than observed effects of changing therapy. We therefore treat this result as exploratory evidence of threshold-dependent utility, not evidence that model-guided treatment adaptation improves patient outcomes.

\subsection{Receptor-Subtype Stratification}
\label{sec:subtype_stratification}

\begin{table}[!t]
\centering
\caption{Receptor-subtype stratification of \corejepa{} five-fold OOF AUROC at each prospective decision point. Values are pooled out-of-fold AUROC; brackets are patient-bootstrap 95\% confidence intervals.}
\label{tab:subtype_stratification}
\scriptsize
\setlength{\tabcolsep}{2.8pt}
\renewcommand{\arraystretch}{1.07}
\begin{tabular}{l r@{\,}l c c c}
\toprule
Subtype & \multicolumn{2}{c}{$n$ (pCR\%)} & T0 & T0--T1 & T0--T2 (95\% CI) \\
\midrule
Overall & 808 & (34.0) & 0.733 & 0.755 & 0.762 \\
HR+/HER2$-$ & 320 & (17.8) & 0.683 & 0.689 & 0.705\,[0.616, 0.784] \\
HER2+ & 201 & (46.8) & 0.712 & 0.741 & 0.731\,[0.658, 0.801] \\
TNBC & 287 & (43.2) & 0.640 & 0.669 & 0.707\,[0.646, 0.765] \\
\bottomrule
\end{tabular}
\end{table}

Table~\ref{tab:subtype_stratification} stratifies the five-fold OOF predictions of \corejepa{} by receptor subtype. At T0--T2, the largest pairwise AUROC difference across subtypes was 0.026, and all pairwise confidence intervals crossed zero, indicating that the overall discrimination was not driven by a single subtype. The overall AUROC of 0.762 exceeded each within-subtype value, reflecting the contribution of cross-subtype differences in baseline pCR prevalence to the aggregate discrimination. Among subtypes, TNBC exhibited the largest longitudinal gain from T0 to T0--T2 ($+0.067$; 95\% CI, 0.010--0.124); this estimate was not corrected for multiple comparisons and is reported as exploratory.

\subsection{Condition Dependence and Response-State Grounding}
\label{sec:condition_grounding}

We examined whether the frozen response state used the clinical and treatment conditions supplied to the transition model. For every outer fold, the fold-specific \corejepa{} checkpoint and validation-selected native response-state \flr{} were retained while the held-out MRI latents and lesion geometry were fixed. Treatment was permuted within HR/HER2 strata, clinical profile was permuted within treatment families, and the joint clinical-treatment context was permuted globally. Each scenario used 100 permutations per fold. This design changes only the conditioning pathway and therefore tests whether context contributes to the organization of the learned state.

At the primary T0--T2 decision point, the observed-context response state achieved $0.7621\pm0.0365$ AUROC. Permuting treatment within biological strata reduced AUROC to $0.7329\pm0.0321$, a mean reduction of $0.0292\pm0.0086$. Permuting clinical profile within treatment family produced a larger reduction to $0.6476\pm0.0210$, and jointly permuting clinical and treatment context reduced AUROC to $0.5951\pm0.0242$ (Fig.~\ref{fig:condition_dependence_5fold}). Every perturbation reduced AUROC in all five folds. The larger effect of clinical-profile permutation is consistent with the cumulative ablation, in which molecular profile and age supplied the largest conditioning gain. The smaller treatment-permutation effect should be interpreted in the context of limited treatment-subtype overlap. These perturbations establish model dependence on the observed conditions, but they neither identify causal treatment effects nor estimate outcomes under unassigned regimens.

Finally, we tested whether the cumulative response-state update retained an observable measure of treatment response. Within each fold, an outcome-free ridge probe mapped the frozen T0-to-T2 state change to log-volume shrinkage, $\log(V_{\mathrm{T0}}/V_{\mathrm{T2}})$. Its regularization strength was selected using the corresponding validation partition, and prediction was restricted to the held-out partition. Across 791 OOF patients, predicted and observed shrinkage had Spearman correlation 0.8623 (95\% CI, 0.8347--0.8865), fold-mean correlation $0.8608\pm0.0500$, and pooled $R^2=0.6744$ (Fig.~\ref{fig:response_state_grounding_5fold}). Median observed shrinkage was 1.9005 among non-pCR patients and 2.7631 among pCR patients ($p=7.03\times10^{-15}$, descriptive Mann--Whitney test). Because lesion-volume change contributes to the imaging-derived response guidance used during pretraining, this analysis verifies that the intended geometric response signal is retained and generalizes to held-out patients; it does not represent independent discovery of an unsupervised volume biomarker.

\section{Discussion}
\label{sec:discussion}

The cumulative ablation shows that patient profile contributes the largest performance gain (image-only JEPA 0.5816 $\to$ 0.7135). This result is consistent with the well-established association between receptor subtype and pCR rates~\cite{cortazar2014,spring2020}: pCR exceeds 40\% in HER2-positive and triple-negative tumors while remaining below 20\% in HR-positive/HER2-negative disease. Knowing the patient's receptor status and associated clinical characteristics therefore provides a strong baseline for pCR discrimination, and the conditioning mechanism allows this prior to shape the learned state trajectory from the outset. The condition-permutation analysis (Fig.~\ref{fig:condition_dependence_5fold}) further confirms this role: permuting clinical profile within treatment families reduced five-fold AUROC from 0.7621 to 0.6476 at T0--T2, and jointly permuting clinical and treatment context reduced it to 0.5951, with every perturbation reducing AUROC in all five folds.

Treatment conditioning yielded no additional marginal AUROC increment in the ablation ($-0.0019$), yet the condition-permutation analysis shows that the transition model does encode treatment information in the frozen response state. Permuting treatment within biological strata reduced T0--T2 AUROC by $0.0292\pm0.0086$ across all five folds (Fig.~\ref{fig:condition_dependence_5fold}). The contrast between the ablation and permutation results reflects the strong correlation between treatment assignment and receptor subtype in I-SPY2, where biological subtype largely determines treatment-arm eligibility. Treatment conditioning therefore serves a representational role, specifying the context under which the next state is predicted, rather than acting as an independent source of pCR discrimination in this cohort. 

\textbf{Limitations.} The image-only JEPA baseline achieved an AUROC of only 0.5816 at T0--T1, and the matched 3-D ResNet and PrefixGRU baselines similarly remained below 0.60 at most decision points. These results indicate that learning discriminative imaging features from the available sample size remains challenging for all evaluated architectures. The explicit geometry pathway and response guidance partially compensate for this by encoding structural priors about lesion morphology, but the reliance on handcrafted geometry descriptors limits the model's capacity to discover novel imaging biomarkers. Improving image-only representation learning under limited longitudinal data is an important direction for future work. Although cross-cohort transfer to I-SPY1 provides preliminary evidence of generalization, independent external validation on larger and more diverse cohorts remains necessary and is currently limited by the availability of longitudinal DCE-MRI datasets with matched clinical data and annotations.

\section{Conclusion}

We presented \corejepa{}, a world-model framework that formulates longitudinal pCR prediction as conditioned next-state prediction rather than direct outcome classification. The model learns a state-transition function from serial DCE-MRI, assigned therapy, and patient clinical profile, producing frozen response states that are linearly probed for pCR prediction at each decision point. Our experiments establish world-model-based representation learning as a promising framework for outcome prediction in oncology and beyond.

\section*{Data and Code Availability}

I-SPY1 and I-SPY2 are de-identified public collections distributed by TCIA \cite{ispy1data,ispy2data}. Code for preprocessing, training, evaluation, and figure generation will be made publicly available upon publication.

\bibliographystyle{IEEEtran}
\bibliography{refs}

\end{document}